\documentclass[letterpaper, 10 pt, conference]{ieeeconf}  %

\IEEEoverridecommandlockouts                              %

\usepackage{graphics} %
\usepackage{epsfig} %
\usepackage{mathptmx} %
\usepackage{times} %

\usepackage{amsmath,amsfonts,amssymb}
\usepackage{array}
\usepackage{booktabs}
\usepackage{cite}
\usepackage{graphicx}
\usepackage{multirow}
\usepackage[caption=false,font=footnotesize]{subfig}
\usepackage{textcomp}
\usepackage[hidelinks]{hyperref}

\title{\LARGE \bf
VLaRL: Augmenting Vision-Language-Action Models with Simulation-Trained Latent-Conditioned Residual RL
}

\author{Namiko Saito$^{1}$, Kinam Kim$^{1,2}$, Heecheol Kim$^{1}$,
Katsushi Ikeuchi$^{1,3}$, and Yasuyuki Matsushita$^{1}$%
\thanks{\raggedright
$^{1}$All authors are with Microsoft Research Asia, Tokyo, Japan.
Corresponding author: Namiko Saito
({\tt\small namikosaito@microsoft.com}).\newline
$^{2}$Kinam Kim is also with KAIST, South Korea.\newline
$^{3}$Katsushi Ikeuchi is also with The University of Tokyo, Japan.}}

\begin{document}

\maketitle
\thispagestyle{empty}
\pagestyle{empty}

\begin{abstract}

Vision-language-action (VLA) models provide broad, instruction-conditioned manipulation behaviors, but their physical execution can remain imprecise during contact-rich interaction.
Residual reinforcement learning (RL) can correct such errors while keeping the VLA frozen, but real-robot RL is costly and safety-critical.
We propose VLA Latent-Conditioned RL (VLaRL), which enables residual RL for frozen VLAs to be trained in simulation and deployed on real robots without real-world RL or online adaptation.
The key challenge is transferring the learned residual policy despite the visual gap between simulation and reality.
Rather than requiring pixel-level visual correspondence, VLaRL uses the VLA's internal vision-language latent representation to condition residual control and as the sim-to-real transfer interface, and learns a lightweight mapper that transforms simulation-derived latents toward the real latent distribution.
Across four contact-rich manipulation tasks and two VLA backbones, VLaRL improves real-world success in all task--backbone combinations, while controlled ablations demonstrate the importance of both latent conditioning and latent alignment for transferring simulation-trained residual control.
The video is available at \url{https://youtu.be/fCkMXTdt1gk}.

\end{abstract}

\section{Introduction}
\label{sec:introduction}
Large-scale Vision-Language-Action (VLA) models have demonstrated broad, language-conditioned behaviors across diverse robot manipulation tasks~\cite{brohan2022rt1,zitkovich2023rt2, ghosh2024octo, kim2024openvla,reuss2025flower, nvidia2025gr00t}.
Their vision-language representations enable a single policy to interpret visual observations and language instructions and generate task-relevant robot actions.
However, successful real-world manipulation also requires precise physical execution, particularly during contact-rich interaction, where small errors in contact location, alignment, or sustained contact can lead to task failure.
A VLA may therefore produce an appropriate nominal behavior while still failing during physical interaction.
Correcting such execution errors through additional imitation learning requires collecting robot demonstration data that capture the desired corrections and updating the base policy.

~%
This motivates us to formulate the improvement of VLA execution as a residual reinforcement learning problem: rather than relearning the complete policy, we keep the VLA frozen and learn only corrective actions for precise physical execution.
Residual RL is well suited to this setting because the pretrained VLA already provides task-relevant nominal behavior, allowing the residual policy to focus on corrections that can be optimized directly from task rewards.
However, acquiring such corrections through RL on physical robots requires repeated interaction and introduces substantial cost and safety concerns.
We therefore aim to train the residual policy entirely through simulation interaction and transfer the learned corrections to the real robot without real-world RL or online adaptation.
\begin{figure}[t]
    \centering
    \includegraphics[width=0.43\textwidth]{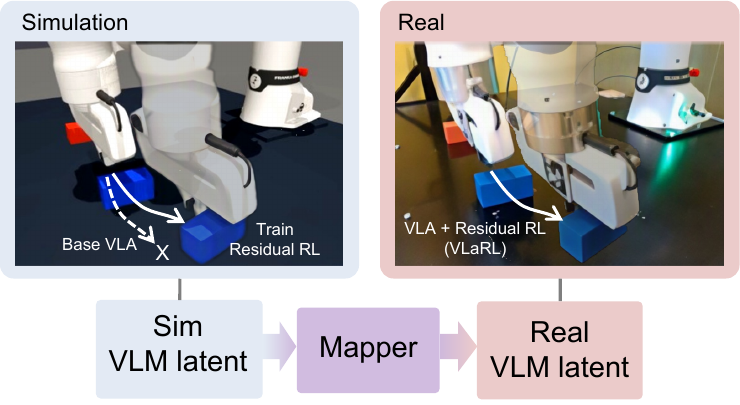}
    \caption{
    Concept of VLaRL. A simulation-trained residual policy locally corrects the nominal behavior of a frozen VLA for precise physical execution. 
    Sim-to-real alignment of the VLM latent representation enables the learned corrections to transfer to the real robot without real-world RL or online adaptation. %
    }
    \vspace{-4mm}
    \label{fig:concept}
\end{figure}
Transferring the residual policy from simulation to reality, however, requires overcoming the visual gap between the two domains.
Existing sim-to-real approaches commonly rely on either visual observations or explicit geometric states as the policy interface.
Image-based policies retain rich scene information but are directly exposed to the visual sim-to-real gap~\cite{tobin2017domain}, whereas geometric representations such as object poses provide a compact interface but require a separate perception pipeline and task-specific state specification~\cite{kim2026objectcentric}.
In this work, we instead explore the internal vision-language representation already computed by the frozen VLA as the interface between simulation-trained residual control and real-world deployment.
Compared with raw visual observations, this representation provides a compact feature space that abstracts visual observations while retaining task-relevant visual and language context.
Reusing this representation therefore allows the residual policy to leverage the VLM's pretrained visual abstraction rather than learning a separate visual representation from simulated images.

However, the VLM representation is not fully invariant to the visual sim-to-real gap: simulated and real observations can still produce different latent distributions.
To make this representation usable as a sim-to-real interface, we learn a lightweight mapper that reduces the resulting latent distribution gap by mapping simulation-derived VLM representations toward the real distribution as shown in Fig.~\ref{fig:concept}.
Building on this aligned interface, we propose \textbf{VLA Latent-Conditioned RL (VLaRL)}, which augments a frozen VLA with a latent-conditioned residual policy trained through reinforcement learning in simulation.
During training, the residual policy learns corrective actions from mapped simulated latents, while at deployment it directly consumes real VLM latents and corrects the nominal VLA action.
This enables residual RL to be trained entirely through simulation interaction and deployed without real-world RL or online adaptation.

We evaluate VLaRL on four contact-rich manipulation tasks---button pressing, block pushing, cup stacking, and drawer closing---using two substantially different VLA backbones, Flower~\cite{reuss2025flower} and GR00T N1.7~\cite{nvidia2025gr00t}.
VLaRL improves real-world success in all eight task--backbone combinations without real-world RL or online adaptation.
Controlled ablations further show that removing the latent mapper substantially degrades real-world performance, highlighting the importance of latent alignment for transferring simulation-trained residual control.

Our main contributions are:
\begin{itemize}
    \item We introduce VLaRL, which augments a frozen VLA with latent-conditioned residual RL for precise physical execution, while preserving VLA's nominal instruction-conditioned behavior.
    \item We use the VLM's internal vision-language latent representation as a sim-to-real interface and learn a lightweight mapper from simulated to real latent distributions, enabling simulation-trained residual corrections to transfer to the real robot without real-world RL or online adaptation.
    \item We validate VLaRL across four contact-rich manipulation tasks and two substantially different VLA backbones, improving real-world success in all eight task--backbone combinations, with controlled ablations demonstrating the importance of both the VLM latent and its sim-to-real alignment.
\end{itemize}
\begin{figure*}[t]
    \centering
    \includegraphics[width=0.94\textwidth]{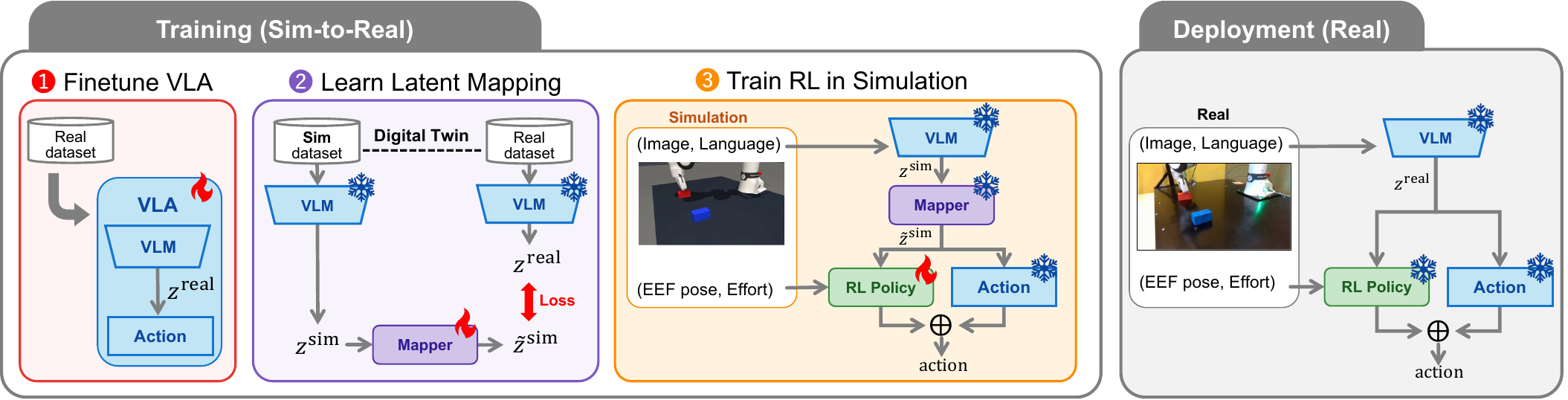}
    \caption{
    Overview of VLaRL and its four-stage workflow: (1) fine-tune the VLA, (2) train the latent mapper, (3) train the residual policy in simulation with the VLA and mapper frozen, and (4) deploy the frozen VLA and residual policy on the real robot. During deployment, real-image latents are supplied directly and the mapper is not used.
    }
    \vspace{-2mm}
    \label{fig:overview}
\end{figure*}

\section{Related Work}
\label{sec:related_work}
\subsection{Vision-Language-Action Models}
Vision-language-action (VLA) models learn general-purpose robot policies from visual observations, language instructions, and robot demonstration data.
RT-1 and RT-2 demonstrated large-scale multi-task robot learning and the transfer of pretrained vision-language representations to robotic control~\cite{brohan2022rt1,zitkovich2023rt2}, while Open X-Embodiment scaled robot learning across datasets and embodiments~\cite{openx2023}.
Recent generalist policies, including OpenVLA~\cite{kim2024openvla}, Octo~\cite{ghosh2024octo}, Flower~\cite{reuss2025flower}, GR00T~\cite{nvidia2025gr00t}, and the $\pi$-series models~\cite{intelligence2026pi07steerablegeneralistrobotic, xu2026rltokenbootstrappingonline}, explore different architectures for mapping vision-language representations to robot actions.

VLaRL builds on these models by keeping the VLA frozen as the nominal controller and learning residual corrections through simulation-based RL.

\subsection{Reinforcement Learning for VLA Refinement}
Residual RL augments an existing controller with additive corrective actions while retaining the base policy as a prior~\cite{silver2018residual,johannink2019residual}, and has been applied to demonstration-based policies~\cite{alakuijala2021residual}, dexterous grasping~\cite{ceola2024resprect}, and sim-to-real correction through human interventions~\cite{jiang2025transic}.
Recent approaches combine RL with imitation or VLA policies in different ways: ResiP trains a state-based residual in simulation and distills the improved behavior into a visual policy~\cite{ankile2024resip}, ResFiT performs residual RL on the physical robot~\cite{ankile2025resfit}, PLD uses residual RL to generate data for subsequent VLA distillation~\cite{xiao2025pld}, and RPD optimizes an RL student initialized from VLA behavior~\cite{julg2025rpd}.
RL Token uses an internal VLM representation as a compact interface for online RL refinement on the physical robot~\cite{xu2026rltokenbootstrappingonline}, whereas object-centric residual RL trains in simulation using estimated 6-DoF object poses as a sim-to-real interface~\cite{kim2026objectcentric}.%

Our VLaRL instead uses the frozen VLA's internal vision-language representation as the residual interface while training the residual entirely in simulation, with latent mapping enabling transfer to the real robot without real-world RL.

\subsection{Sim-to-Real Transfer and Representation Alignment}
Sim-to-real robot learning addresses discrepancies in observations and dynamics~\cite{zhao2020sim2real}, commonly through domain randomization~\cite{tobin2017domain,peng2018dynamics,chebotar2019closing,ramos2019bayessim,ma2024dreureka} or compact geometric representations such as 6-DoF object pose~\cite{wen2024foundationpose}.
Real-to-sim approaches instead reconstruct or generate simulation environments from real observations, as in RialTo~\cite{villasevil2024rialto} and Digital Cousins~\cite{dai2025digital}.

VLaRL similarly uses reconstructed task configurations for simulation training but transfers residual control through the pretrained VLM representation rather than requiring pixel-level visual correspondence or an explicit object-state interface at deployment.
Because pretrained representations are not necessarily invariant to the visual sim-to-real gap, VLaRL maps simulation-derived VLM latents toward the real latent distribution during residual training, allowing the deployed residual policy to directly consume real VLM latents.

\section{Method}
\label{sec:method}
\subsection{Overview}
\label{sec:method_overview}
VLaRL augments a frozen VLA with a residual policy trained entirely in simulation. 
We treat the frozen VLA as two functional components: a vision-language module that encodes visual observations and language instructions into an internal latent representation, and an action head that generates nominal robot actions from this representation.
VLaRL reuses the intermediate vision-language representation as the input to residual control while retaining the original action head as the nominal controller.

Given visual observation $o_t$ at time $t$ and language instruction $\ell$, the vision-language module first produces a token matrix
\begin{equation}
    Z_t = E_{\mathrm{VLM}}(o_t, \ell),
    \qquad
    Z_t \in \mathbb{R}^{S \times D},
    \label{eq:vla_tokens}
\end{equation}
where $E_{\mathrm{VLM}}$ denotes the vision-language module within the frozen VLA, and $S$ and $D$ denote the number and dimension of tokens, respectively.
Conditioned on $Z_t$, the VLA action head $\pi_{\mathrm{act}}$ then produces the nominal action 
\begin{equation}
    a_t^{\mathrm{VLA}}
    \sim
    \pi_{\mathrm{act}}
    \left(
        \cdot \mid Z_t
    \right),
    \label{eq:vla_action}
\end{equation}
which is the action generated by the frozen VLA before residual correction.~%

For residual control, the token matrix is reduced to a $D$-dimensional latent vector $z_t \in \mathbb{R}^{D}$ and used as the interface from the VLM to residual control.
The residual policy predicts a corrective action
\begin{equation}
    a_t^{\mathrm{RL}}
    =
    \pi_{\mathrm{RL}}
    \left(
        z_t,
        a_t^{\mathrm{VLA}},
        s_t,
        f_t
    \right),
    \label{eq:residual_policy}
\end{equation}
where $z_t$ denotes the fixed-dimensional VLM representation supplied to the residual policy, $s_t$ is the robot proprioceptive state, and $f_t$ is the measured Cartesian force on the wrist.%
The executed action is
\begin{equation}
    a_t
    =
    a_t^{\mathrm{VLA}}
    +
    \alpha a_t^{\mathrm{RL}},
    \label{eq:residual_action}
\end{equation}
where $\alpha$ is a fixed residual scaling factor shared across tasks.

Training this residual policy entirely in simulation introduces a visual sim-to-real gap: simulated and real observations can produce different VLM representation distributions even for similar task states.
As shown in Fig.~\ref{fig:overview}, VLaRL addresses this gap through four steps: (1) fine-tuning and freezing the VLA and reconstructing the corresponding simulation, (2) learning a mapper that transforms simulation-derived VLM representations toward the real latent distribution, (3) training the residual policy entirely in simulation using the mapped representation, and (4) deploying the learned residual policy directly on real VLM representations without the mapper, real-world RL, or online adaptation.

\subsection{Step 1: VLA Fine-Tuning and Digital Twin Reconstruction}
\label{sec:vla_preparation}
For each task, we collect real-robot demonstration data and fine-tune the VLA.
The fine-tuned VLA is then frozen throughout the remaining stages.
Using the same demonstrations, we reconstruct approximately corresponding task configurations in simulation as digital twins, which provide the sim--real observations used for latent mapping and the simulation environment for residual RL.

\subsection{Step 2: Sim-to-Real VLM Latent Mapping}
\label{sec:latent_mapping}
Simulated and real observations differ visually even for similar task states, producing a gap in their VLM representations. 
Because the VLA is fine-tuned on real demonstrations, we learn a mapper that transforms simulation-derived representations toward the real latent distribution.
The mapper serves two roles: reducing the representation shift seen by the VLA action head in simulation, and aligning the residual-policy inputs between simulation training and real-world deployment.

Let $Z_t^{s}, Z_t^{r} \in \mathbb{R}^{S\times D}$ denote the VLM token representations extracted from simulated and real observations, respectively.
The mapper $M_\theta$ operates on the simulated token matrix before pooling:
\begin{equation}
    \widetilde{Z}_t^{s}
    =
    M_\theta
    \left(
        Z_t^{s}
    \right).
    \label{eq:latent_mapper}
\end{equation}
Because the reconstructed sim--real trajectories provide approximate rather than exact frame-level correspondence, we train $M_\theta$ using optimal-transport-based distribution alignment.
For simulated sample $i$ and real sample $j$, we define the transport cost as
\begin{equation}
    C_{ij}
    =
    \frac{
        \left\|
            M_\theta(Z_i^{s}) - Z_j^{r}
        \right\|_F^2
    }{SD}
    +
    \lambda_{\tau}
    \left(
        \tau_i^{s} - \tau_j^{r}
    \right)^2,
    \label{eq:ot_cost}
\end{equation}
where $\tau$ denotes normalized trajectory progress and $\lambda_\tau$ weights the trajectory-progress cost.
Using this cost, we optimize the mapper with entropy-regularized optimal transport, which enables efficient computation of the transport plan using Sinkhorn iterations~\cite{cuturi2013sinkhorn}.~%

The mapped representation $\widetilde{Z}_t^s$ is supplied to the frozen VLA action head and mean-pooled for the residual policy:
\begin{equation}
    \widetilde{z}_t^{s}
    =
    \operatorname{MeanPool}
    \left(
        \widetilde{Z}_t^{s}
    \right).
    \label{eq:mapped_latent}
\end{equation}
The mapper is frozen during residual RL.

\subsection{Step 3: Simulation-Trained Residual Reinforcement Learning}
\label{sec:residual_rl}
With the mapper fixed, we train the residual policy entirely through interaction in simulation using TD3~\cite{fujimoto2018addressing}.
At each step, the frozen VLA action head produces the nominal action from the mapped simulated token representation $\widetilde{Z}_t^s$, while its mean-pooled representation $\widetilde{z}_t^s$ is supplied to the residual actor.
The residual actor therefore receives $\widetilde{z}_t^{s}$ together with the nominal VLA action, proprioceptive state $s_t$, and Cartesian force $f_t$, following Eq.~\eqref{eq:residual_policy}.

We use an asymmetric actor--critic formulation in which the critic additionally receives privileged task state available in simulation, while the actor uses only observations available at real-world deployment.
To stabilize learning around the demonstrated behavior, we additionally construct residual targets from the demonstrations:
\begin{equation}
    a_t^{\mathrm{RL,demo}}
    =
    \frac{
        a_t^{\mathrm{demo}}
        -
        a_t^{\mathrm{VLA}}
    }{\alpha},
    \label{eq:demo_residual}
\end{equation}
and use them as an auxiliary regularization term during residual-policy training.
The VLA and latent mapper remain frozen throughout RL training; only the residual policy is optimized.

\subsection{Step 4: Real-Robot Deployment}
\label{sec:deployment}

At deployment, the frozen VLA produces the nominal action $a_t^{\mathrm{VLA}}$ and real token representation $Z_t^{r}$ from the real observation and language instruction.
The real representation is directly mean-pooled as
\begin{equation}
    z_t^{r}
    =
    \operatorname{MeanPool}
    \left(
        Z_t^{r}
    \right),
    \label{eq:real_latent}
\end{equation}
and supplied to the residual policy together with the nominal VLA action, proprioceptive state $s_t$, and Cartesian force $f_t$.
Because the residual policy is trained on simulated representations mapped toward the real latent distribution, the mapper is not used at deployment.
All models remain frozen, and no real-world RL or online adaptation is performed.

\section{Experimental Setup}
\label{sec:experimental_setup}
\subsection{Robot Setup and Manipulation Tasks}
\label{sec:robot_setup}
Experiments are conducted using a Franka Research 3 robot in both the real world and MuJoCo simulation~\cite{todorov2012mujoco}.
With the real robot, we collected 32 demonstrations per task by teleoperation with a 3D mouse.
To reconstruct the corresponding digital-twin configurations, we use FoundationPose~\cite{wen2024foundationpose} to reconstruct the object position and the environment.
The same demonstrations are used for VLA finetuning, sim--real latent mapping, and demonstration-derived residual targets.

\begin{figure*}[t]
    \centering
    \subfloat[Button pressing]{\includegraphics[width=0.225\textwidth]{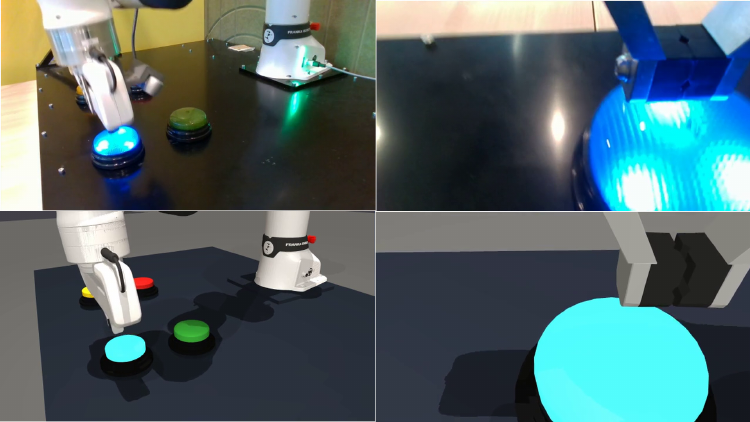}}\hfill
    \subfloat[Block pushing]{\includegraphics[width=0.225\textwidth]{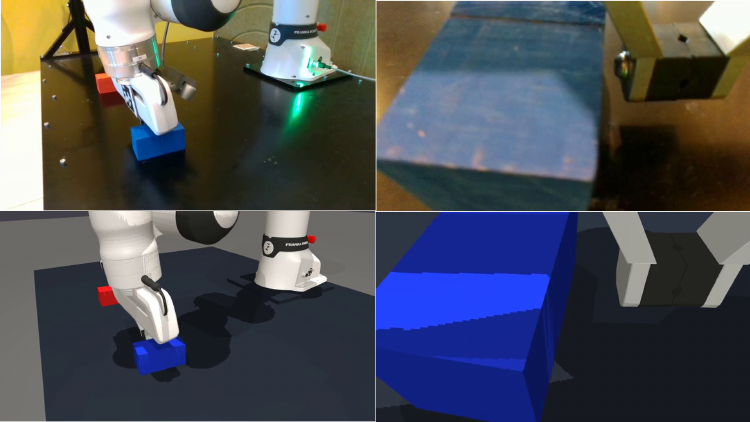}}\hfill
    \subfloat[Cup stacking]{\includegraphics[width=0.225\textwidth]{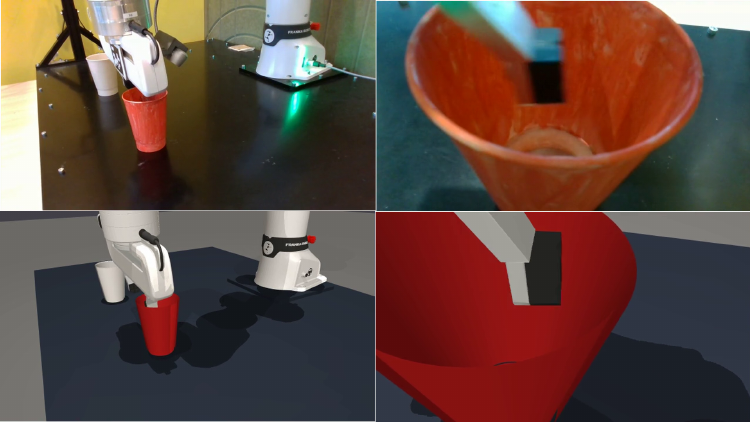}}\hfill
    \subfloat[Drawer closing]{\includegraphics[width=0.225\textwidth]{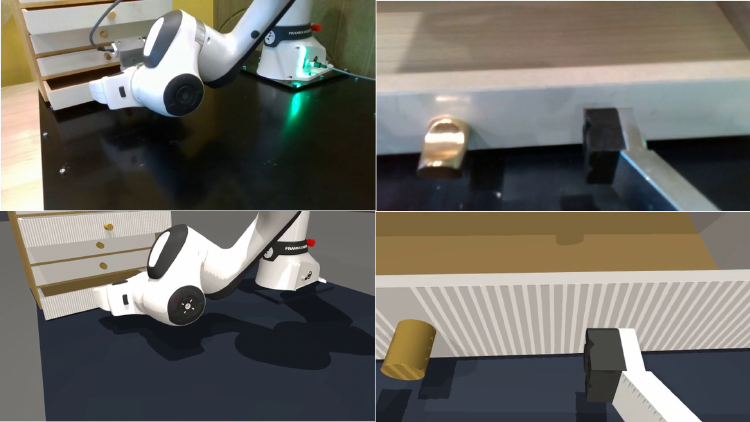}}
    \caption{The four manipulation tasks and their digital-twin observations.
    Within each task panel, the top row shows the real left- and hand-camera images, and the bottom row shows the corresponding simulated left- and hand-camera images extracted at the same time step.}
    \label{fig:tasks}
\end{figure*}

We evaluate four contact-rich manipulation tasks: button pressing, block pushing, cup stacking, and drawer closing, covering precise contact, sustained contact, spatial alignment, and articulated contact, respectively.
The tasks and corresponding digital-twin observations are shown in Fig.~\ref{fig:tasks}, with language instructions and success criteria summarized in Table~\ref{tab:tasks}.

We additionally evaluate unseen-object transfer for block pushing and cup stacking.
The pushing policy trained on rectangular blocks is tested on novel star- and triangle-shaped objects, while the stacking policy is tested on novel sky-blue and green cups under both language-specified role assignments.
No additional residual RL or real-world adaptation is performed for these objects.
\subsection{Training Details}
\label{sec:training_details}
To evaluate the applicability of our framework across different VLA
architectures, we used two VLA backbones, Flower~\cite{reuss2025flower} and GR00T N1.7~\cite{nvidia2025gr00t}.
For Flower, we extract language-conditioned tokens from its Florence-2-based vision-language module, yielding a 1024-dimensional representation after mean pooling; for GR00T N1.7, we use its language-conditioned image tokens, yielding a 2048-dimensional representation.
For both backbones, the normalized Cartesian VLA action is 7-dimensional, comprising three translations, three rotations, and one gripper command.

The latent mapper is a four-layer Transformer encoder with 16 attention heads and feed-forward dimension twice the VLA token dimension, operating on the token sequence before mean pooling.
The mapper uses a learnable residual connection whose scaling factor is initialized to zero, such that it initially behaves as an identity transformation.
The mapper is trained with Sinkhorn divergence using $\lambda_\tau=0.5$, entropic regularization $\varepsilon=0.05$, and 50 Sinkhorn iterations.

The residual policy receives the normalized Cartesian VLA action, the end-effector position and roll--pitch--yaw orientation, gripper state, and 3-dimensional Cartesian force.
Residual policies are trained for 60,000 environment steps using TD3~\cite{fujimoto2018addressing}.
The actor and twin critics are four-layer ReLU MLPs with hidden width 1024; the actor outputs a 7-dimensional $\tanh$ residual action and is initialized to produce near-zero corrections.
We use batch size 256, discount factor $\gamma=0.99$, Polyak coefficient $\tau=0.005$, and residual scale $\alpha=0.10$.
The actor is additionally regularized using the demonstration-derived residual targets described in Sec.~\ref{sec:residual_rl} and the residual-action magnitude, while the critics receive privileged task state available only in simulation.
The environment reward combines task progress with a terminal success bonus. 
The task-progress rewards encourage target contact for button pressing, sustained displacement with limited rotation for block pushing, reaching and aligning the instructed cups for cup stacking, and reducing the instructed drawer opening for drawer closing.

\begin{table*}[t]
    \centering
    \caption{Evaluation tasks and success criteria.}
    \label{tab:tasks}
    \begin{tabular}{llll}
        \hline
        \textbf{Task} &
        \textbf{Local-control challenge} &
        \textbf{Instruction} &
        \textbf{Success criterion} \\
        \hline

        Press button &
        Precise target contact &
        \textit{``Press the red / blue / yellow / green button''} &
        Beep + light \\

        Push block &
        Sustained contact &
        \textit{``Push the blue / red block''} &
        Displacement $>10$ cm \\

        Stack cup &
        Grasp + spatial alignment &
        \textit{``Put the red / white cup inside the white / red cup''} &
        Overlap $>4$ cm \\

        Close drawer &
        Articulated contact &
        \textit{``Close the upper / bottom drawer'' }  &
        Opening $<1$ cm \\

        \hline
    \end{tabular}
\end{table*}
\subsection{Evaluation Protocol}
\label{sec:evaluation_protocol}

Initial object positions are randomized by $\pm3$ cm and orientations by $\pm15^\circ$; for drawer closing, the drawer opening is randomized by $\pm3$ cm.
Simulation results are reported over three random seeds, and each real-world condition is evaluated over 40 trials.
All models remain frozen during real-world evaluation, with no real-world RL or online adaptation.

We compare the frozen VLA with VLaRL and conduct two controlled ablations using Flower.
\textit{w/o VLM latent} removes the latent while retaining the nominal VLA action, proprioception, and force, whereas \textit{w/o Mapper} trains the residual policy on unaligned simulated VLM latents.
Both ablations follow the same evaluation protocol and success criteria as VLaRL.

\section{Results}
\label{sec:results}

\subsection{Simulation-Trained Residual Control Improves Real-World Execution}
\label{sec:main_results}

\begin{figure*}[t]
    \centering
    \includegraphics[width=0.84\textwidth]{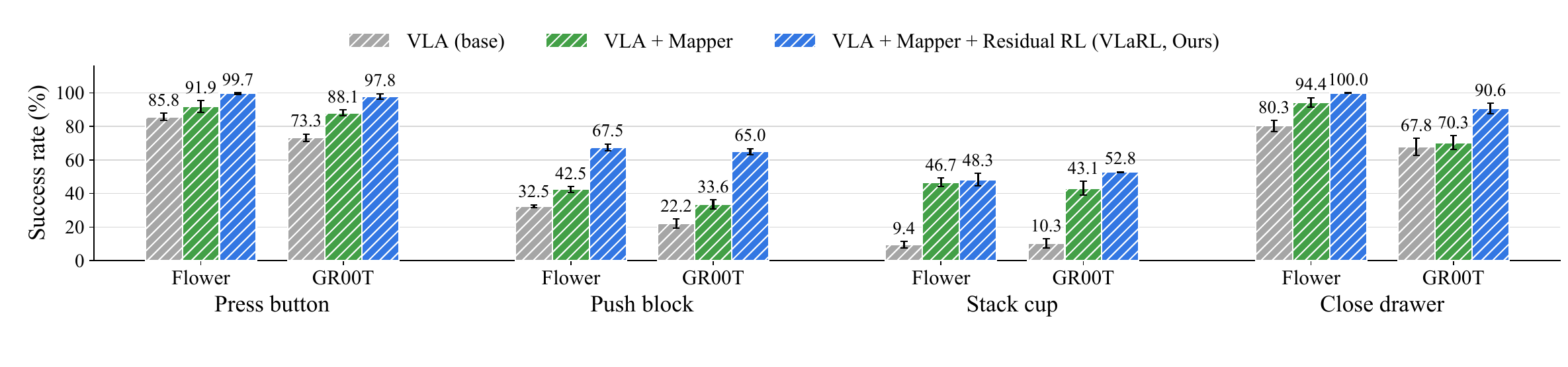}
    \vspace{-8mm}
    \caption{Simulation success rates across the four task categories and two VLA backbones.
    VLA + Mapper + Residual RL denotes VLaRL (Ours).
    Error bars report the variation over three seeds.}
    \label{fig:sim-success}
\end{figure*}

\begin{figure*}[t]
    \centering
    \includegraphics[width=0.84\textwidth]{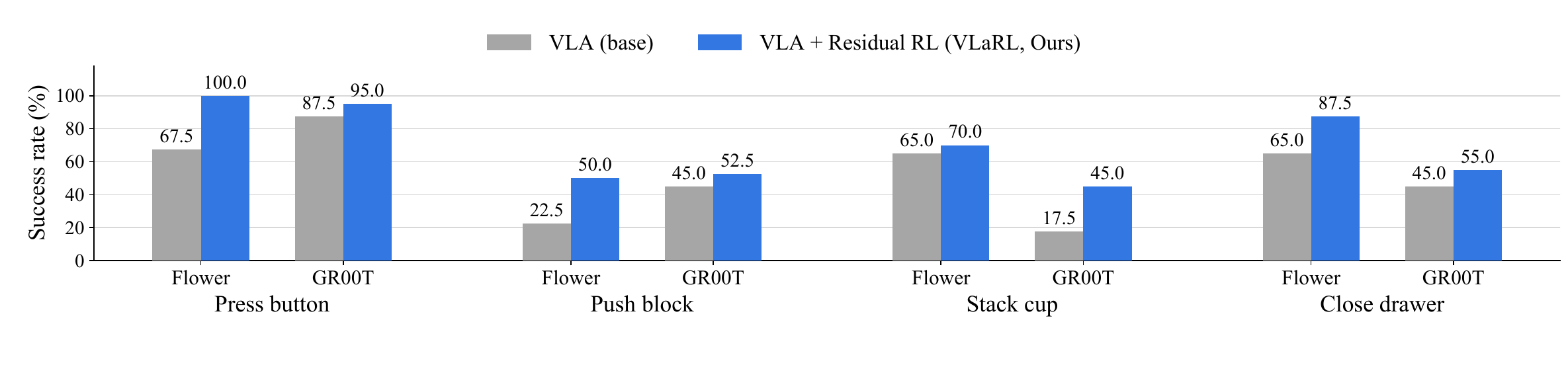}
    \vspace{-8mm}
    \caption{Real-robot success rates over 40 trials for each task and VLA backbone.
    VLA + Residual RL denotes VLaRL (Ours), which improves the frozen base VLA in all eight evaluated combinations.}
    \label{fig:real-success}
\end{figure*}

Figures~\ref{fig:sim-success} and~\ref{fig:real-success} summarize the simulation and real-robot results across four tasks and two VLA backbones.
In simulation, applying the mapper improves the nominal VLA success across all task--backbone combinations, consistent with reducing the representation shift encountered when the real-data-finetuned VLA is applied to simulated observations.
Adding residual RL further improves success across all combinations; for example, Flower block-pushing success increases from 32.5\% with the base VLA to 42.5\% with the mapper and 67.5\% with VLaRL.

\begin{figure*}[t]
    \centering
    \includegraphics[width=0.94\textwidth]{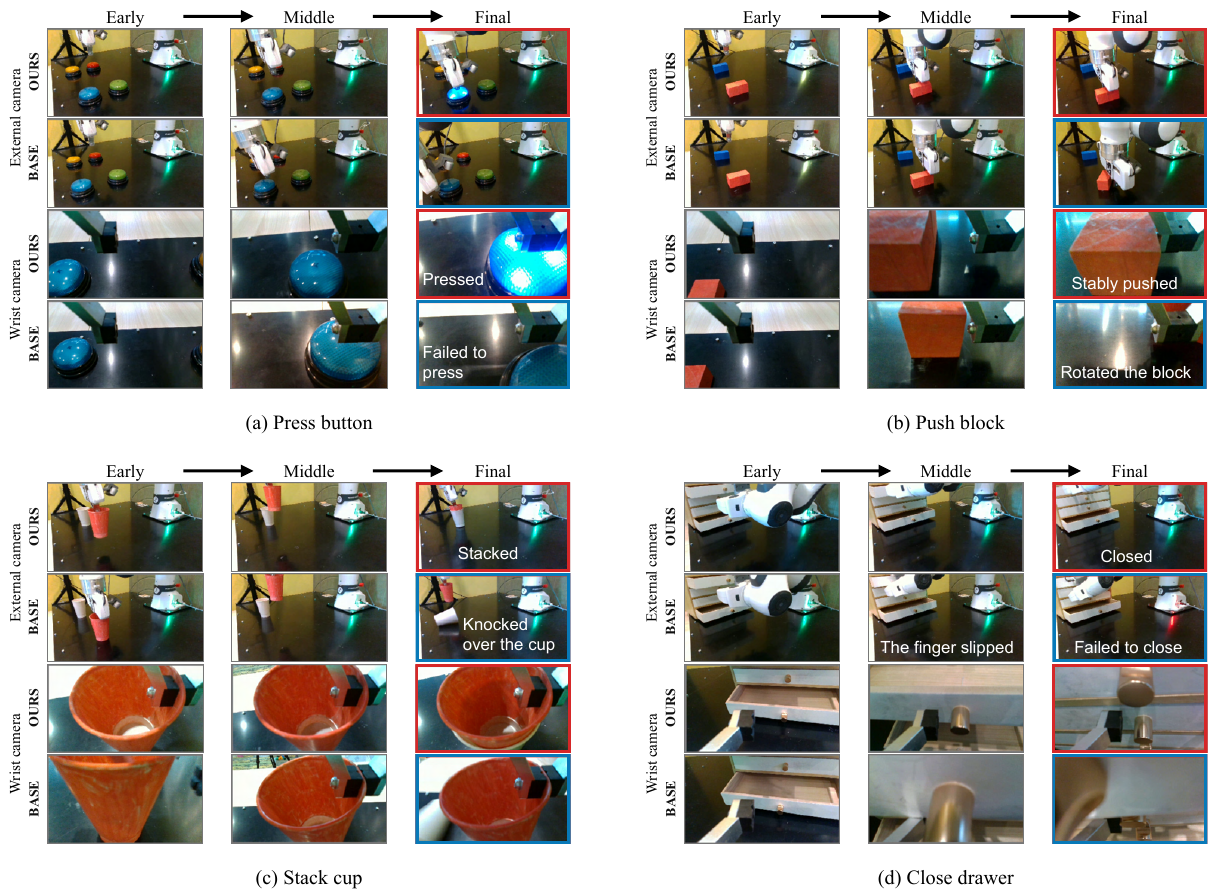}
    \caption{
    Representative Flower real-robot rollout progressions for the base VLA and VLaRL across the four tasks.
    Columns show temporal progression from early to final states; rows show the base and proposed methods for the left and hand cameras.}
    \label{fig:task-progressions}
\end{figure*}

The simulation-trained residual policies transfer to the real robot without real-world RL or online adaptation, with VLaRL improving success over the base VLA in all eight task--backbone combinations.
The gains are particularly pronounced for Flower button pressing (67.5\% to 100.0\%), Flower block pushing (22.5\% to 50.0\%), and GR00T N1.7 cup stacking (17.5\% to 45.0\%), covering precise target contact, sustained contact, and spatial alignment, respectively.
Representative Flower real-world rollouts in Fig.~\ref{fig:task-progressions} show that the frozen VLA produces task-relevant nominal behavior but can fail during local physical execution, while our VLaRL with the residual correction enables successful completion in the illustrated examples.

Simulation and real-world performance do not always follow the same trend.
For example, Flower achieves substantially higher base success on cup stacking in reality than in simulation, and the residual provides only a small additional real-world improvement.
This discrepancy indicates that latent alignment does not eliminate other sim-to-real differences, such as contact dynamics and the distribution of VLA actions.

\subsection{Ablation Analysis}
\label{sec:ablations}

\begin{figure}[t]
    \centering
    \includegraphics[width=0.95\columnwidth]{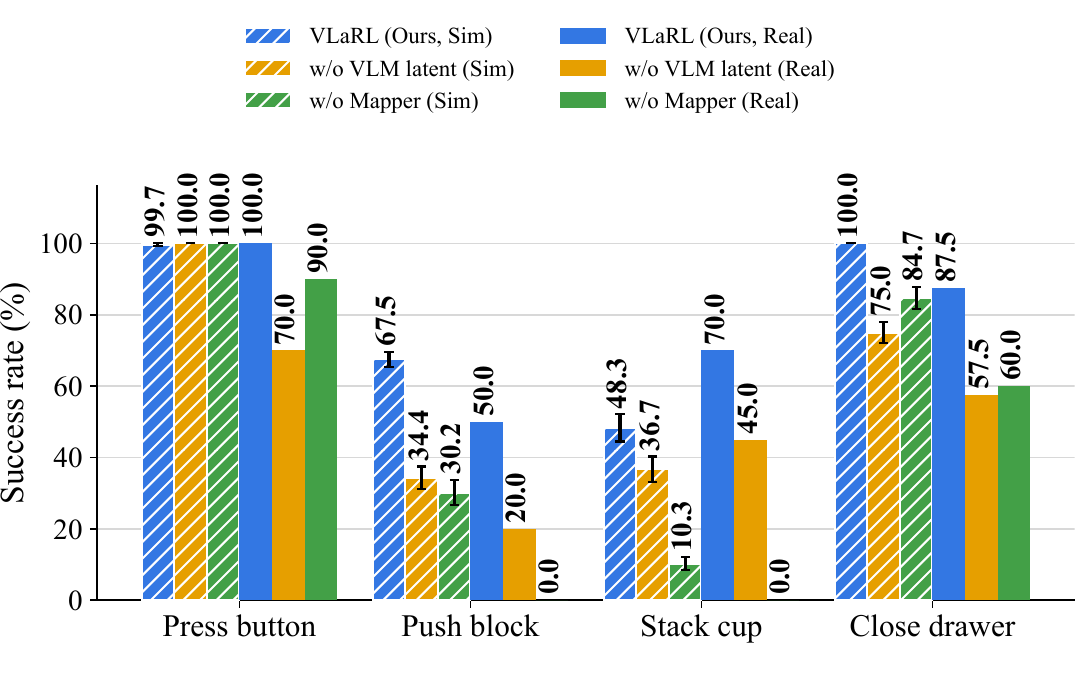}
    \caption{Flower ablations of the VLM latent input and sim-to-real latent mapper.
    ``VLaRL (Ours)'' is the complete system; ``w/o VLM latent'' retains the nominal VLA action, proprioception, and force, while ``w/o Mapper'' uses raw simulated VLM latents.
    Hatched bars denote simulation and solid bars denote real-robot evaluation.
    Simulation error bars report variation over three seeds; real-robot results use 40 trials per task.}
    \vspace{-2mm}
    \label{fig:component-ablation}
\end{figure}

We next evaluate the role of latent alignment in real-world residual transfer and the contribution of VLM-latent conditioning using Flower.
As shown in Fig.~\ref{fig:component-ablation}, removing the mapper substantially degrades real-world performance.
In particular, block pushing and cup stacking fail in all 40 real-robot trials without the mapper, compared with 50.0\% and 70.0\% success with the full VLaRL, respectively.
This result shows that residual policies trained on unaligned simulated VLM representations do not directly transfer to real representations, highlighting the importance of latent mapping for sim-to-real residual learning.

Removing the VLM latent while retaining the nominal VLA action, proprioception, and force also degrades real-world performance on all four tasks, including button pressing (100.0\% to 70.0\%) and block pushing (50.0\% to 20.0\%).
This indicates that the VLM latent provides information useful for residual control beyond the nominal action and robot state.

\begin{figure}[]
    \centering
    \includegraphics[width=\columnwidth]{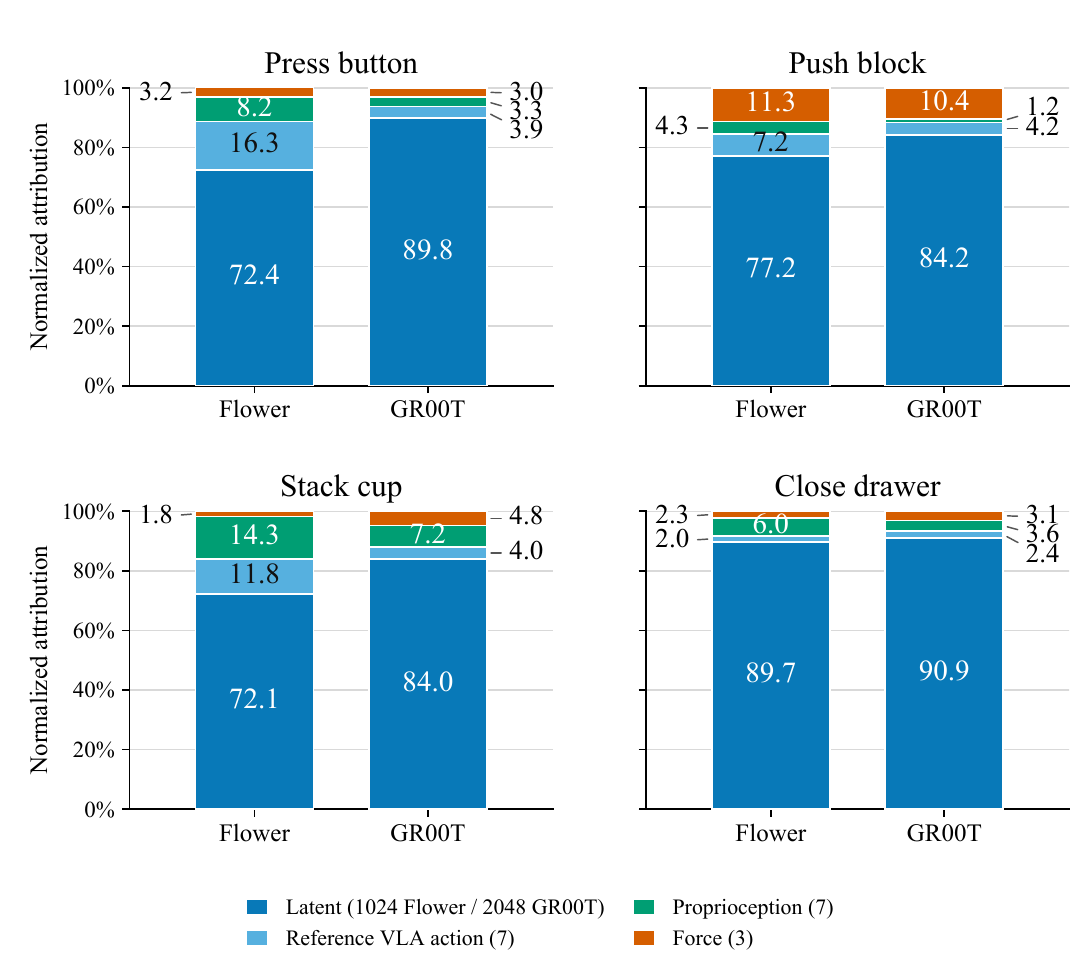}
    \captionsetup{font=small}
    \caption{Integrated-gradients attribution of the residual-action norm to the VLM latent, reference VLA action, proprioception, and force.
    Absolute group attributions are normalized to sum to 100\%.}
    \vspace{-2mm}
    \label{fig:ig}
\end{figure}

We further analyze the residual policy using integrated gradients~\cite{sundararajan2017axiomatic}.
As shown in Fig.~\ref{fig:ig}, the VLM latent receives the largest attribution to the residual-action norm across all four tasks and both VLA backbones, ranging from 72.1\% to 89.7\% for Flower and from 84.0\% to 90.9\% for GR00T N1.7. 
The relative contribution of the physical inputs varies across tasks: force attribution rises for block pushing, which requires sustained contact, while proprioception becomes relatively more important for cup stacking, which requires precise position adjustment. 
These task-dependent patterns suggest that the residual policy combines the VLM representation with physical feedback according to the local-control requirements of each task. 
Together with the controlled latent ablation, this analysis indicates that the learned residual policy makes substantial use of the VLM representation while also incorporating complementary physical information.

\subsection{Unseen-Object Transfer}
\label{sec:unseen_results}

\begin{figure}[]
    \centering
    \subfloat[Unseen cups]{\includegraphics[width=0.46\columnwidth]{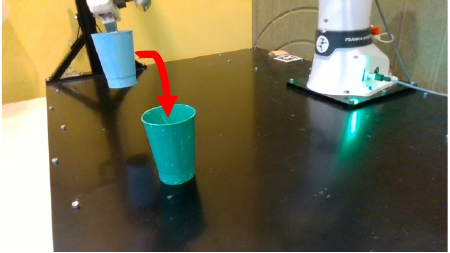}}\hfill
    \subfloat[Unseen shapes]{\includegraphics[width=0.46\columnwidth]{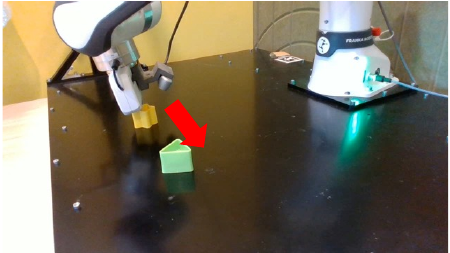}}
    \caption{Transfer to objects not used during residual-policy training.
    The cup experiment changes appearance and language-assigned roles, while the pushing experiment changes object geometry.}
    \label{fig:unseen-objects}
\end{figure}

Although the mapper and residual policy are trained on a limited set of objects, we further evaluate VLaRL on unseen objects as shown in Fig.~\ref{fig:unseen-objects}, to assess whether the residual corrections preserve the generalization of the frozen VLA.
For block pushing, the unseen star- and triangle-shaped objects are evaluated across balanced target-object and left--right placement conditions.
For cup stacking, the unseen sky-blue and green cups are evaluated across balanced language-specified source--target roles and left--right placements.
Without additional residual RL or real-world adaptation, VLaRL improves real-world success from 40.0\% (16/40) to 57.5\% (23/40) for block pushing and from 55.0\% (22/40) to 62.5\% (25/40) for cup stacking.
These results show that the learned residual corrections remain effective under unseen object variation while preserving instruction-dependent target and object-role selection from the frozen VLA.

\section{Discussion}
\label{sec:discussion}

\subsection{Why Latent-Conditioned Residual RL?}
\label{sec:discussion_latent}

Residual control provides a way to improve precise physical execution without retraining or replacing the underlying VLA.
The frozen VLA remains responsible for instruction-conditioned nominal behavior, while RL provides bounded local corrections for contact, alignment, and interaction errors.
Our results across Flower and GR00T N1.7 show that this interface can be applied to different VLA designs, while the unseen-object results show that residual correction can improve execution under object variation while retaining the instruction-dependent target and object-role selection of the frozen VLA.

In VLaRL, conditioning the residual policy on the VLM latent allows it to reuse a representation derived from both visual observations and language instructions.
Unlike the nominal VLA action alone, the VLM latent exposes the visual
and instruction context from which the action was generated.
This is particularly relevant when the same scene requires different corrections depending on the instructed target or manipulation role.
The latent ablation and integrated-gradients analysis show that the residual policy makes substantial use of this representation beyond the nominal VLA action, proprioception, and force.
Explicit geometric states such as object poses provide an alternative compact and interpretable interface, but require task-relevant objects and state variables to be specified and estimated in advance.
VLM latents instead provide a richer interface containing appearance, semantic, and instruction context without requiring an object-pose estimator at deployment.

\subsection{Simulation-Only Residual Learning}
\label{sec:discussion_transfer}

A central motivation of VLaRL is to avoid costly real-world RL interaction by learning residual corrections entirely in simulation.
The challenge is that visual differences between simulation and reality propagate into the VLM representation, so a residual policy trained directly on simulated latents may not transfer to real latents.
VLaRL addresses this problem by mapping simulation-derived representations toward the real latent distribution before residual RL training.
The mapper ablation shows that this alignment is important for real-world transfer, enabling simulation-trained residual corrections to be deployed on the real robot without real-world RL or online adaptation.
This provides a practical way to exploit large-scale simulation interaction for improving the physical execution of frozen VLA policies.

\subsection{Limitations and Future Directions}
\label{sec:limitations}

Several limitations remain.
First, training the latent mapper requires approximately corresponding sim--real trajectories constructed through task-specific digital-twin reconstruction.
Although this process is performed only during training, reducing or eliminating the need for such paired data would make the approach easier to scale.

Second, the mapper addresses differences in VLM representations arising from the visual sim-to-real gap, but does not explicitly address discrepancies in contact dynamics, friction, compliance, or other physical properties.
These remaining differences may explain discrepancies between simulation and real-world performance.

Third, the mapper is specific to each VLA backbone, and the residual RL is currently trained separately for each task category.
Learning transferable latent alignment and shared residual policies across backbones and manipulation skills is an important direction toward more general residual control.

Finally, our unseen-object evaluation covers a limited range of object variations and does not establish open-world generalization.
Nevertheless, the improvements on unseen objects suggest that residual learning can improve local execution while preserving the VLA's ability to generate instruction-dependent nominal behavior for unseen targets.


\section{Conclusion}
\label{sec:conclusion}

We presented VLaRL, which augments a frozen VLA with latent-conditioned residual RL for precise physical execution.
VLaRL reuses the VLA's internal vision-language representation as the residual-control interface and learns a sim-to-real mapper that transforms simulation-derived latents toward the real latent distribution, enabling residual RL to be trained entirely in simulation and deployed without real-world RL or online adaptation.
Across four contact-rich manipulation tasks and two different VLA backbones, Flower and GR00T N1.7, VLaRL consistently improved real-world success, demonstrating its applicability across different VLA architectures, while ablations demonstrated the importance of both latent conditioning and
sim-to-real latent mapping.
The residual policies also improved execution on unseen objects while retaining the instruction-dependent nominal behavior of the frozen VLA.
These results demonstrate a practical approach to combining the generalizable behavior of pretrained VLAs with simulation-trained residual control for precise real-world manipulation.

\bibliographystyle{IEEEtran}
\bibliography{references}

\end{document}